\documentclass{article}

\usepackage{microtype}
\usepackage{graphicx}
\usepackage{subfigure}
\usepackage{booktabs} % for professional tables

\usepackage[hyphens]{url}
\usepackage{hyperref}
\usepackage{amsfonts}
\usepackage{amsmath}
\usepackage{cleveref}
\usepackage{mathtools}

\usepackage[symbol]{footmisc}

\DeclareMathOperator*{\argmin}{argmin} % thin space, limits underneath in displays
\DeclareMathOperator*{\argmax}{argmax} % thin space, limits underneath in displays
\graphicspath{{img/}}

\usepackage[accepted]{icml2020}

\icmltitlerunning{Interpretable Counterfactual Explanations Guided by Prototypes}

\begin{document}

\twocolumn[
\icmltitle{Interpretable Counterfactual Explanations Guided by Prototypes}

% It is OKAY to include author information, even for blind
% submissions: the style file will automatically remove it for you
% unless you've provided the [accepted] option to the icml2020
% package.

% List of affiliations: The first argument should be a (short)
% identifier you will use later to specify author affiliations
% Academic affiliations should list Department, University, City, Region, Country
% Industry affiliations should list Company, City, Region, Country

% You can specify symbols, otherwise they are numbered in order.
% Ideally, you should not use this facility. Affiliations will be numbered
% in order of appearance and this is the preferred way.
%\icmlsetsymbol{equal}{*}

\begin{icmlauthorlist}
    \icmlauthor{Arnaud Van Looveren}{seldon}
    \icmlauthor{Janis Klaise}{seldon}
\end{icmlauthorlist}

\icmlaffiliation{seldon}{Seldon Technologies Ltd, London, United Kingdom}

\icmlcorrespondingauthor{Arnaud Van Looveren}{avl@seldon.io}
\icmlcorrespondingauthor{Janis Klaise}{jk@seldon.io}

% You may provide any keywords that you
% find helpful for describing your paper; these are used to populate
% the "keywords" metadata in the PDF but will not be shown in the document
\icmlkeywords{Machine Learning, ICML}

\vskip 0.3in
]

% this must go after the closing bracket ] following \twocolumn[ ...

% This command actually creates the footnote in the first column
% listing the affiliations and the copyright notice.
% The command takes one argument, which is text to display at the start of the footnote.
% The \icmlEqualContribution command is standard text for equal contribution.
% Remove it (just {}) if you do not need this facility.

\printAffiliationsAndNotice{}  % leave blank if no need to mention equal contribution
%\printAffiliationsAndNotice{\icmlEqualContribution} % otherwise use the standard text.

\begin{abstract}
    We propose a fast, model agnostic method for finding interpretable counterfactual explanations of classifier predictions by using class prototypes. We show that class prototypes, obtained using either an encoder or through class specific k-d trees, significantly speed up the search for counterfactual instances and result in more interpretable explanations. We quantitatively evaluate interpretability of the generated counterfactuals to illustrate the effectiveness of our method on an image and tabular dataset, respectively MNIST and Breast Cancer Wisconsin (Diagnostic). Additionally, we propose a principled approach to handle categorical variables and illustrate our method on the Adult (Census) dataset. Our method also eliminates the computational bottleneck that arises because of numerical gradient evaluation for \textit{black box} models.\footnote{An open source implementation of the algorithm can be found
    at \url{https://github.com/SeldonIO/alibi}.}
\end{abstract}

\section{Introduction}
Humans often think about how they can alter the outcome of a situation. \textit{What do I need to change for the bank to approve my loan?} or \textit{Which symptoms would lead to a different medical diagnosis?} are common examples. This form of counterfactual reasoning comes natural to us and explains how to arrive at a desired outcome in an interpretable manner. Moreover, examples of counterfactual instances resulting in a different outcome can give powerful insights of what is important to the underlying decision process, making it a compelling method to explain predictions of machine learning models (\Cref{fig:cf_examples}).

In the context of predictive models, given a test instance and the model's prediction, a counterfactual instance describes the necessary change in input features that alter the prediction to a predefined output \cite{molnar2019}. For classification models the predefined output can be any target class or prediction probability distribution. Counterfactual instances can then be found by iteratively perturbing the input features of the test instance until the desired prediction is reached. In practice, the counterfactual search is posed as an optimization problem---we want to minimize an objective function which encodes desirable properties of the counterfactual instance with respect to the perturbations. The key insight of this formulation is the need to design an objective function that allows us to generate high quality counterfactual instances. A counterfactual instance $x_{\text{cf}}$ should have the following desirable properties:

\begin{figure}[t]
    \centering
    \includegraphics[width=0.98\columnwidth]{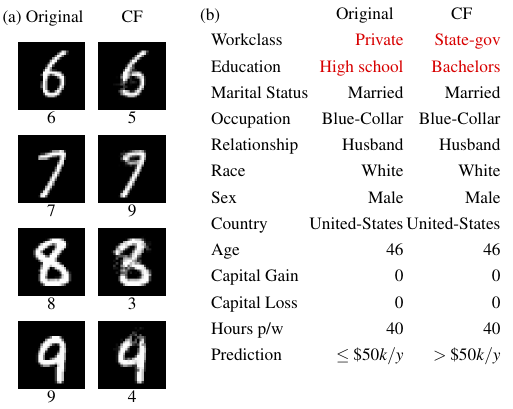}
    \caption{(a) Examples of original and counterfactual instances on the MNIST dataset along with predictions of a CNN model. (b) A counterfactual instance on the Adult (Census) dataset highlighting the feature changes required to alter the prediction of an NN model.}
    \label{fig:cf_examples}
\end{figure}

\begin{enumerate}
    \item The model prediction on $x_{\text{cf}}$ needs to be close to the predefined output.
    \item The perturbation $\delta$ changing the original instance $x_{0}$ into $x_{\text{cf}}=x_{0}+\delta$ should be sparse.
    \item The counterfactual $x_{\text{cf}}$ needs to be interpretable. We consider an instance $x_{\text{cf}}$ interpretable if it lies close to the model's training data distribution. This definition does not only apply to the overall data set, but importantly also to the \textit{training instances that belong to the counterfactual class}. Let us illustrate this with an intuitive example. Assume we are predicting house prices with features including the square footage and the number of bedrooms. Our house is valued below \pounds500,000 and we would like to know what needs to change about the house in order to increase the valuation above \pounds500,000. By simply increasing the number of bedrooms and leaving the other features unchanged, the model predicts that our \textit{counterfactual house} is now worth more than \pounds500,000. This sparse counterfactual instance lies fairly close to the overall training distribution since only one feature value was changed. The counterfactual is however out-of-distribution with regards to the subset of houses in the training data valued above \pounds500,000 because other relevant features like the square footage still resemble a typical house valued below \pounds500,000. As a result, we do not consider this counterfactual to be very interpretable. We show in the experiments that there is often a trade-off between sparsity and interpretability.
    \item The counterfactual instance $x_\text{cf}$ needs to be found fast enough to ensure it can be used in a real life setting.
\end{enumerate}
An overly simplistic objective function may return instances which satisfy properties \text{1.} and \text{2.}, but where the perturbations are not interpretable with respect to the counterfactual class.

In this paper we propose using class prototypes in the objective function to guide the perturbations quickly towards an interpretable counterfactual. The prototypes also allow us to remove computational bottlenecks from the optimization process which occur due to numerical gradient calculation for black box models. In addition, we propose two novel metrics to quantify interpretability which provide a principled benchmark for evaluating interpretability at the instance level. We show empirically that prototypes improve the quality of counterfactual instances on both image (MNIST) and tabular (Wisconsin Breast Cancer) datasets. Finally, we propose using pairwise distance measures between categories of categorical variables to define meaningful perturbations for such variables and illustrate the effectiveness of the method on the Adult (Census) dataset.

\section{Related Work}

Counterfactual instances---synthetic instances of data engineered from real instances to change the prediction of a machine learning model---have been suggested as a way of explaining individual predictions of a model as an alternative to feature attribution methods such as LIME \cite{ribeiro2016} or SHAP \cite{lundberg2017}.

\citet{wachter2018} generate counterfactuals by minimizing an objective function which sums the squared difference between the predictions on the perturbed instance and the desired outcome, and a scaled \text{$L_{1}$} norm of the perturbations. \citet{thibault2018} find counterfactuals through a heuristic search procedure by growing spheres around the instance to be explained. The above methods do not take local, class specific interpretability into account. Furthermore, for black box models the number of prediction calls during the search process grows proportionally to either the dimensionality of the feature space \cite{wachter2018} or the number of sampled observations \cite{thibault2018,dhurandhar2019model}, which can result in a computational bottleneck. \citet{dhurandhar2018,dhurandhar2019model} propose the framework of \textit{Contrastive Explanations} which find the minimal number of features that need to be changed/unchanged to keep/change a prediction.

A key contributions of this paper is the use of prototypes to guide the counterfactual search process. \citet{beenkim2016criticism,gurumoorthy2017protodash} use prototypes as example-based explanations to improve the interpretability of complex datasets. Besides improving interpretability, prototypes have a broad range of applications like clustering \cite{kaufmanl1987clustering}, classification \cite{bien2011,TAKIGAWA2009101}, and few-shot learning \cite{snell2017}. If we have access to an encoder \cite{rumelhart1986}, we follow the approach of \cite{snell2017} who define a class prototype as the mean encoding of the instances which belong to that class. In the absence of an encoder, we find prototypes through class specific k-d trees \cite{Bentley1975}.

To judge the quality of the counterfactuals we introduce two novel metrics which focus on local interpretability with respect to the training data distribution. This is different from \cite{dhurandhar2017tip} who define an interpretability metric relative to a target model. \citet{beenkim2016criticism} on the other hand quantify interpretability through a human pilot study measuring the accuracy and efficiency of the humans on a predictive task. \citet{luss2019generating} also highlight the importance of good local data representations in order to generate high quality explanations.

Another contribution of this paper is a principled approach to handling categorical variables during the counterfactual generation process. Some previously proposed solutions are either computationally expensive \cite{wachter2018} or do not take relationships between categories into account \cite{dhurandhar2019model, Mothilal_2020}. We propose using pairwise distance measures to define embeddings of categorical variables into numerical space which allows us to define meaningful perturbations when generating counterfactuals.

\section{Methodology}
\subsection{Background}
\begin{figure}[ht]
    \centering
    \includegraphics[width=0.9\columnwidth]{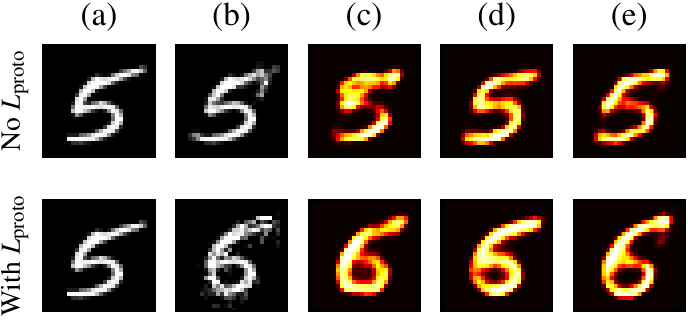}
    \caption{First row: (a) original instance and (b) uninterpretable counterfactual $3$. (c), (d) and (e) are reconstructions of (b) with respectively $\text{AE}_{3}$, $\text{AE}_{5}$ and $\text{AE}$. Second row: (a) original instance and (b) interpretable counterfactual $6$. (c), (d) and (e) are reconstructions of (b) with respectively $\text{AE}_{6}$, $\text{AE}_{5}$ and $\text{AE}$.}
    \label{fig:bad_cf}
\end{figure}

The following section outlines how the prototype loss term is constructed and why it improves the convergence speed and interpretability. Finding a counterfactual instance $x_{\text{cf}} = x_{0} + \delta$, with both $x_{\text{cf}}$ and $x_{0}$ $\in$ $\mathcal{X}\subseteq\mathbb{R}^D$ where $\mathcal{X}$ represents the $D$-dimensional feature space, implies optimizing an objective function of the following form:

\begin{equation}\label{eq:main_form}
    \min_{\delta} c \cdot f_{\kappa}(x_{0},\delta) + f_{\text{dist}}(\delta).
\end{equation}
$f_{\kappa}(x_{0},\delta)$ encourages the predicted class $i$ of the perturbed instance $x_{\text{cf}}$ to be different than the predicted class $t_{0}$ of the original instance $x_{0}$. Similar to \cite{dhurandhar2018}, we define this loss term as:

\begin{equation}
    \begin{aligned}
    L_{\text{pred}}&\coloneqq f_{\kappa}(x_{0},\delta)\\
    &= \max([f_{\text{pred}}(x_{0} + \delta )]_{t_{0}} - \max_{i \neq t_{0}}[f_{\text{pred}}(x_{0} + \delta)]_{i}, - \kappa),
    \end{aligned}
\end{equation}
where $[f_{\text{pred}}(x_{0} + \delta )]_{i}$ is the $i$-th class prediction probability, and $\kappa \geq 0$ caps the divergence between $[f_{\text{pred}}(x_{0} + \delta )]_{t_{0}}$ and $[f_{\text{pred}}(x_{0} + \delta )]_{i}$. The term $f_{\text{dist}}(\delta)$ minimizes the distance between $x_{0}$ and $x_{\text{cf}}$ with the aim to generate sparse counterfactuals. We use an elastic net regularizer \cite{zou2005en}:

\begin{equation}
    f_{\text{dist}}(\delta) = \beta \cdot \| \delta \|_{1} + \| \delta \|_{2}^2 = \beta \cdot L_{1} + L_{2}.
\end{equation}
While the objective function \eqref{eq:main_form} is able to generate counterfactual instances, it does not address a number of issues:

\begin{enumerate}
    \item $x_{\text{cf}}$ does not necessarily respect the training data manifold, resulting in out-of-distribution counterfactual instances. Often a trade off needs to be made between sparsity and interpretability of $x_{\text{cf}}$.
    \item The scaling parameter $c$ of $f_{\kappa}(x_{0},\delta)$ needs to be set within the appropriate range before a potential counterfactual instance is found. Finding a good range can be time consuming.
\end{enumerate}
\citet{dhurandhar2018} aim to address the first issue by adding in an additional loss term $L_{\text{AE}}$ which represents the $L_{2}$ reconstruction error of $x_{cf}$ evaluated by an autoencoder AE which is fit on the training set:

\begin{equation}
    L_{\text{AE}} = \gamma \cdot \| x_{0} + \delta - \text{AE}(x_{0} + \delta) \|_{2}^2.
\end{equation}
The loss $L$ to be minimized now becomes:

\begin{equation}\label{eq:loss_ae}
    L = c \cdot L_{\text{pred}} + \beta \cdot L_{1} + L_{2} +  L_{\text{AE}}.
\end{equation}
The autoencoder loss term $L_{\text{AE}}$ penalizes out-of-distribution counterfactual instances, but does not take the data distribution for each prediction class $i$ into account. This can lead to sparse but uninterpretable counterfactuals, as illustrated by \Cref{fig:bad_cf}. The first row of \Cref{fig:bad_cf}(b) shows a sparse counterfactual $3$ generated from the original $5$ using loss function \eqref{eq:loss_ae}. Both visual inspection and reconstruction of the counterfactual instance using \text{AE} in \Cref{fig:bad_cf}(e) make clear however that the counterfactual lies closer to the distribution of a $5$ and is not interpretable as a $3$. The second row adds a prototype loss term to the objective function, leading to a less sparse but more interpretable counterfactual $6$.

The $L_{\text{AE}}$ loss term also does not consistently speed up the counterfactual search process since it imposes a penalty on the distance between the proposed $x_{\text{cf}}$ and its reconstruction by the autoencoder without explicitly guiding $x_{\text{cf}}$ towards an interpretable solution. We address these issues by introducing an additional loss term, $L_{\text{proto}}$.

\subsection{Prototype loss term}

By adding in a prototype loss term $L_{\text{proto}}$, we obtain the following objective function:

\begin{equation}
    L = c \cdot L_{\text{pred}} + \beta \cdot L_{1} + L_{2} +  L_{\text{AE}} + L_{\text{proto}},
\end{equation}
where $L_{\text{AE}}$ becomes optional. The aim of $L_{\text{proto}}$ is twofold:

\begin{enumerate}
    \item Guide the perturbations $\delta$ towards an interpretable counterfactual $x_{\text{cf}}$ which falls in the distribution of counterfactual class $i$.
    \item Speed up the counterfactual search process without too much hyperparameter tuning.
\end{enumerate}
To define the prototype for each class, we can reuse the encoder part of the autoencoder from $L_{\text{AE}}$. The encoder $\text{ENC}(x)$ projects $x \in\mathcal{X}$ onto an $E$-dimensional latent space $\mathbb{R}^E$. We also need a representative, unlabeled sample of the training dataset. First the predictive model is called to label the dataset with the classes predicted by the model. Then for each class $i$ we encode the instances belonging to that class and order them by increasing $L_{2}$ distance to $\text{ENC}(x_{0})$. Similar to \cite{snell2017}, the class prototype is defined as the average encoding over the $K$ nearest instances in the latent space with the same class label:

\begin{equation}
    \mathrm{proto}_{i} \coloneqq \frac{1}{K} \sum_{k=1}^{K} \text{ENC}(x_{k}^{i})
\end{equation}
for the ordered $\lbrace x_{k}^{i}\rbrace_{k=1}^{K}$ in class $i$. It is important to note that the prototype is defined in the latent space, not the original feature space.

The Euclidean distance is part of a class of distance functions called \textit{Bregman divergences}. If we consider that the encoded instances belonging to class $i$ define a cluster for $i$, then $\mathrm{proto}_{i}$ equals the cluster mean. For Bregman divergences the cluster mean yields the minimal distance to the points in the cluster \cite{Banerjee05clusteringwith}. Since we use the Euclidean distance to find the closest class to $x_{0}$, $\mathrm{proto}_{i}$ is a suitable class representation in the latent space. When generating a counterfactual instance for $x_{0}$, we first find the nearest prototype $\mathrm{proto}_{j}$ of class $j \neq t_{0}$ to the encoding of $x_{0}$:

\begin{equation}
    j = \argmin_{i \neq t_{0}} \| \text{ENC}(x_{0}) - \mathrm{proto}_{i} \|_{2}.
\end{equation}
The prototype loss $L_{\text{proto}}$ can now be defined as:

\begin{equation}
    L_{\text{proto}} = \theta \cdot \| \text{ENC}(x_{0} + \delta) - \mathrm{proto}_{j} \|_{2}^2,
\end{equation}
where $\text{ENC}(x_{0} + \delta)$ is the encoding of the perturbed instance. As a result, $L_{\text{proto}}$ explicitly guides the perturbations towards the nearest prototype $\mathrm{proto}_{j \neq t_{0}}$, speeding up the counterfactual search process towards the average encoding of class $j$. This leads to more interpretable counterfactuals as illustrated by the experiments. \Cref{alg:enc} summarizes this approach.

\begin{algorithm}[t]
    \caption{Counterfactual search with encoded prototypes}
    \label{alg:enc}
    \begin{algorithmic}[1]
        \STATE\textbf{Parameters:} $\beta, \theta$ (required) and $c, \kappa$ and $\gamma$ (optional)
        \STATE \textbf{Inputs:} AE (optional) and ENC models. A sample $X=\lbrace x_1,\dots,x_n\rbrace$ from training set. Instance to explain $x_{0}$.
        \STATE Label $X$ and $x_0$ using the prediction function $f_{\text{pred}}$:

        $X^{i} \leftarrow \lbrace x\in X \mid \argmax f_{\text{pred}}(x) = i \rbrace$ for each class $i$
        $t_0 \leftarrow \argmax f_{\text{pred}}(x_0)$
        \STATE Define prototypes for each class $i$:

        $\mathrm{proto}_{i} \leftarrow \tfrac{1}{K} \sum_{k=1}^{K} \text{ENC}(x_{k}^{i}) $ for $ x_{k}^{i} \in X^i$ where $x_{k}^{i}$ is ordered by increasing $\| \text{ENC}(x_{0}) - \text{ENC}(x_{k}^{i}) \|_{2}$ and $\,K \leq \vert X^i\vert$
        
        \STATE Find nearest prototype $j$ to instance $x_{0}$ but different from original class $t_{0}$:
        
        $j \leftarrow \argmin_{i \neq t_{0}} \| \text{ENC}(x_{0}) - \mathrm{proto}_{i} \|_{2}$.
        
        \STATE Optimize the objective function:

        $\delta^{*} \leftarrow \argmin_{\delta \in \mathcal{X}} c \cdot L_{\text{pred}} + \beta \cdot L_{1} + L_{2} + L_{\text{AE}} + L_{\text{proto}}$ where $L_{\text{proto}} = \theta \cdot \| \text{ENC}(x_{0} + \delta) - \mathrm{proto}_{j} \|_{2}^2$.
        \STATE \textbf{Return} $x_{\text{cf}} = x_{0} + \delta^{*}$
    \end{algorithmic}
\end{algorithm}

\begin{algorithm}[!ht]
    \caption{Counterfactual search with k-d trees}
    \label{alg:kd}
    \begin{algorithmic}[1]
            \STATE \textbf{Parameters:} $\beta, \theta, k$ (required) and $c, \kappa$ (optional)
            \STATE \textbf{Input:} A sample $X=\lbrace x_1,\dots,x_n\rbrace$ from training set. Instance to explain $x_{0}$.
            \STATE Label $X$ and $x_0$ using the prediction function $f_{\text{pred}}$:
            
            $X^{i} \leftarrow \lbrace x\in X \mid \argmax f_{\text{pred}}(x) = i \rbrace$ for each class $i$
            $t_0 \leftarrow \argmax f_{\text{pred}}(x_0)$

            \STATE Build separate k-d trees for each class $i$ using $X_i$

            \STATE Find nearest prototype $j$ to instance $x_{0}$ but different from original class $t_{0}$:
            
            $j \leftarrow \argmin_{i \neq t_{0}} \| x_{0} - x_{i,k} \|_{2}$ where $x_{i,k}$ is the $k$-th nearest item to $x_{0}$ in the k-d tree of class $i$.
            
            $\mathrm{proto}_{j} \leftarrow x_{j,k}$
            \STATE Optimize the objective function:

            $\delta^{*} \leftarrow \argmin_{\delta \in \mathcal{X}} c \cdot L_{\text{pred}} + \beta \cdot L_{1} + L_{2} + L_{\text{proto}}$ where $L_{\text{proto}} = \theta \cdot \|x_{0} + \delta - \mathrm{proto}_{j} \|_{2}^2$.

            \STATE \textbf{Return} $x_{\text{cf}} = x_{0} + \delta^{*}$
        \end{algorithmic}
\end{algorithm}

\subsection{Using k-d trees as class representations}

If we do not have a trained encoder available, we can build class representations using k-d trees \cite{Bentley1975}. After labeling the representative training set by calling the predictive model, we can represent each class $i$ by a separate k-d tree built using the instances with class label $i$. This approach is similar to \cite{jiang2018} who use class specific k-d trees to measure the agreement between a classifier and a modified nearest neighbour classifier on test instances. For each k-d tree $j \neq t_{0}$, we compute the Euclidean distance between $x_{0}$ and the $k$-nearest item in the tree $x_{j, k}$. The closest $x_{j, k}$ across all classes $j \neq t_{0}$ becomes the class prototype $\mathrm{proto}_{j}$. Note that we are now working in the original feature space. The loss term $L_{\text{proto}}$ is equal to:

\begin{equation}
    L_{\text{proto}} = \theta \cdot \| x_{0} + \delta - \mathrm{proto}_{j} \|_{2}^2.
\end{equation}
\Cref{alg:kd} outlines the k-d trees approach.

\subsection{Categorical variables}
Creating meaningful perturbations for categorical data is not straightforward as the very concept of perturbing an input feature implies some notion of rank and distance between the values a variable can take. We approach this by inferring pairwise distances between categories of a categorical variable based on either model predictions (Modified Value Distance Metric) \cite{cost1993} or the context provided by the other variables in the dataset (Association-Based Distance Metric) \cite{le2005}. We then apply multidimensional scaling \cite{borg2005mds} to project the inferred distances into one-dimensional Euclidean space, which allows us to perform perturbations in this space. After applying a perturbation in this space, we map the resulting number back to the closest category before evaluating the classifier's prediction.

\subsection{Removing $L_{\text{\normalfont pred}}$}

In the absence of $L_{\text{proto}}$, only $L_{\text{pred}}$ encourages the perturbed instance to predict class $i \neq t_{0}$. In the case of black box models where we only have access to the model's prediction function, $L_{\text{pred}}$ can become a computational bottleneck. This means that for neural networks, we can no longer take advantage of automatic differentiation and need to evaluate the gradients numerically. Let us express the gradient of $L_{\text{pred}}$ with respect to the input features $x$ as follows:

\begin{equation}
    \frac{\partial L_{\text{pred}}}{\partial x} = \frac{\partial f_{\kappa}(x)}{\partial x} = \frac{\partial f_{\kappa}(x)}{\partial f_{\text{pred}}} \frac{\partial f_{\text{pred}}}{\partial x},
\end{equation}
where $f_{\text{pred}}$ represents the model's prediction function. The numerical gradient approximation for $f_{\text{pred}}$ with respect to input feature $k$ can be written as:

\begin{equation}
    \frac{\partial f_{\text{pred}}}{\partial x_{k}} \approx \frac{f_{\text{pred}}(x + \epsilon_{k}) - f_{\text{pred}}(x - \epsilon_{k})}{2 \epsilon},
\end{equation}
where $\epsilon_{k}$ is a perturbation with the same dimension as $x$ and taking value $\epsilon$ for feature $k$ and $0$ otherwise. As a result, the prediction function needs to be evaluated twice for each feature per gradient step just to compute $\tfrac{\partial f_{\text{pred}}}{\partial x_{k}}$. For a $28\times 28$ MNIST image, this translates into a batch of $28 \cdot 28 \cdot 2=1568$ prediction function calls. Eliminating $L_{\text{pred}}$ would therefore speed up the counterfactual search process significantly. By using the prototypes to guide the counterfactuals, we can remove $L_{\text{pred}}$ and only call the prediction function once per gradient update on the perturbed instance to check whether the predicted class $i$ of $x_{0} + \delta$ is different from $t_{0}$. This eliminates the computational bottleneck while ensuring that the perturbed instance moves towards an interpretable counterfactual $x_{\text{cf}}$ of class $i \neq t_{0}$.

\subsection{FISTA optimization}

Like \cite{dhurandhar2018}, we optimize our objective function by applying a fast iterative shrinkage-thresholding algorithm (FISTA) \cite{beck2009} where the solution space for the output $x_{\text{cf}} = x_{0} +  \delta$ is restricted to $\mathcal{X}$. The optimization algorithm iteratively updates $\delta$ with momentum for $N$ optimization steps. It also strips out the $\beta \cdot L_{1}$ regularization term from the objective function and instead shrinks perturbations $| \delta_{k} | < \beta$ for feature $k$ to $0$. The optimal counterfactual is defined as $x_{\text{cf}} = x_{0} + \delta^{n^{*}}$ where $n^{*} = \argmin_{n \in {1, ..., N}} \beta \cdot \| \delta^{n} \|_{1} + \| \delta^{n} \|_{2}^{2}$ and the predicted class on $x_{\text{cf}}$ is $i \neq t_{0}$.

\section{Experiments}

The experiments are conducted on an image and tabular dataset. The first experiment on the MNIST handwritten digit dataset \cite{lecun2010mnist} makes use of an autoencoder to define and construct prototypes. The second experiment uses the Breast Cancer Wisconsin (Diagnostic) dataset \cite{Dua2019}. The latter dataset has lower dimensionality so we find the prototypes using k-d trees. Finally, we illustrate our approach for handling categorical data on the Adult (Census) dataset \cite{Dua2019}.

\subsection{Evaluation}

The counterfactuals are evaluated on their interpretability, sparsity and speed of the search process. The sparsity is evaluated using the elastic net loss term $\text{EN}(\delta) = \beta \cdot \| \delta \|_{1} + \| \delta \|_{2}^2$ while the speed is measured by the time and the number of gradient updates required until a satisfactory counterfactual  $x_{\text{cf}}$ is found. We define a satisfactory counterfactual as the optimal counterfactual found using FISTA for a fixed value of $c$ for which counterfactual instances exist.

In order to evaluate interpretability, we introduce two interpretability metrics IM1 and IM2. Let $\text{AE}_{i}$ and $\text{AE}_{t_{0}}$ be autoencoders trained specifically on instances of classes $i$ and $t_{0}$, respectively. Then IM1 measures the ratio between the reconstruction errors of $x_{\text{cf}}$ using $\text{AE}_{i}$ and $\text{AE}_{t_{0}}$:

\begin{equation}
    \text{IM1}(\text{AE}_i, \text{AE}_{t_0}, x_{\text{cf}}) \coloneqq \frac{\| x_{0} + \delta - \text{AE}_{i}(x_{0} + \delta) \|_{2}^2}{\| x_{0} + \delta - \text{AE}_{t_{0}}(x_{0} + \delta) \|_{2}^2 + \epsilon}.
\end{equation}
A lower value for IM1 means that $x_{\text{cf}}$ can be better reconstructed by the autoencoder which has only seen instances of the counterfactual class $i$ than by the autoencoder trained on the original class $t_{0}$. This implies that $x_{\text{cf}}$ lies closer to the data manifold of counterfactual class $i$ compared to $t_{0}$, which is considered to be more interpretable.

The second metric IM2 compares how similar the reconstructed counterfactual instances are when using $\text{AE}_{i}$ and an autoencoder trained on all classes, $\text{AE}$. We scale IM2 by the $L_{1}$ norm of $x_{\text{cf}}$ to make the metric comparable across classes:
\begin{equation}
    \text{IM2}(\text{AE}_i, \text{AE}, x_{\text{cf}}) \coloneqq \frac{\| \text{AE}_{i}(x_{0} + \delta) - \text{AE}(x_{0} + \delta) \|_{2}^2}{\| x_{0} + \delta \|_{1} + \epsilon}.
\end{equation}
A low value of IM2 means that the reconstructed instances of $x_{\text{cf}}$ are very similar when using either $\text{AE}_{i}$ or $\text{AE}$. As a result, the data distribution of the counterfactual class $i$ describes $x_{\text{cf}}$ as good as the distribution over all classes. This implies that the counterfactual is interpretable. \Cref{fig:bad_cf} illustrates the intuition behind IM1 and IM2. 

The uninterpretable counterfactual $3$ ($x_{\text{cf,1}}$) in the first row of \Cref{fig:bad_cf}(b) has an IM1 value of 1.81 compared to 1.04 for $x_{\text{cf,2}}$ in the second row because the reconstruction of $x_{\text{cf,1}}$ by $\text{AE}_{5}$ in \Cref{fig:bad_cf}(d) is better than by $\text{AE}_{3}$ in \Cref{fig:bad_cf}(c). The IM2 value of $x_{\text{cf,1}}$ is higher as well---0.15 compared to 0.12 for $x_{\text{cf,2}}$)---since the reconstruction by $\text{AE}$ in \Cref{fig:bad_cf}(e) yields a clear instance of the original class $5$.

Finally, for MNIST we apply a multiple model comparison test based on the maximum mean discrepancy \cite{lim2019} to evaluate the relative interpretability of counterfactuals generated by each method.

\subsection{Handwritten digits}
The first experiment is conducted on the MNIST dataset. The experiment analyzes the impact of $L_{\text{proto}}$ on the counterfactual search process with an encoder defining the prototypes for $K$ equal to 5. We further investigate the importance of the $L_{\text{AE}}$ and $L_{\text{pred}}$ loss terms in the presence of $L_{\text{proto}}$. We evaluate and compare counterfactuals obtained by using the following loss functions:
\begin{equation}\label{eq:mnist_losses}
    \begin{aligned}
        A &= c \cdot L_{\text{pred}} + \beta \cdot L_{1} + L_{2}\\
        B &= c \cdot L_{\text{pred}} + \beta \cdot L_{1} + L_{2} + L_{\text{AE}}\\
        C &= c \cdot L_{\text{pred}} + \beta \cdot L_{1} + L_{2} + L_{\text{proto}}\\
        D &= c \cdot L_{\text{pred}} + \beta \cdot L_{1} + L_{2} + L_{\text{AE}} + L_{\text{proto}}\\
        E &= \beta \cdot L_{1} + L_{2} + L_{\text{proto}}\\
        F &= \beta \cdot L_{1} + L_{2} + L_{\text{AE}} + L_{\text{proto}}
    \end{aligned}
\end{equation}

For each of the ten classes, we randomly sample 50 numbers from the test set and find counterfactual instances for 3 different random seeds per sample. This brings the total number of counterfactuals to 1,500 per loss function.

The model used to classify the digits is a convolutional neural network with 2 convolution layers, each followed by a max-pooling layer. The output of the second pooling layer is flattened and fed into a fully connected layer followed by a softmax output layer over the 10 possible classes. For objective functions $B$ to $F$, the experiment also uses a trained autoencoder for the $L_{\text{AE}}$ and $L_{\text{proto}}$ loss terms. The autoencoder has 3 convolution layers in the encoder and 3 deconvolution layers in the decoder. Full details of the classifier and autoencoder, as well as the hyperparameter values used can be found in the supplementary material.
\subsubsection{Results}
\Cref{tb:table_mnist} summarizes the findings for the speed and interpretability measures.
\begin{figure*}[t]
    \centering
    \includegraphics[width=0.95\textwidth]{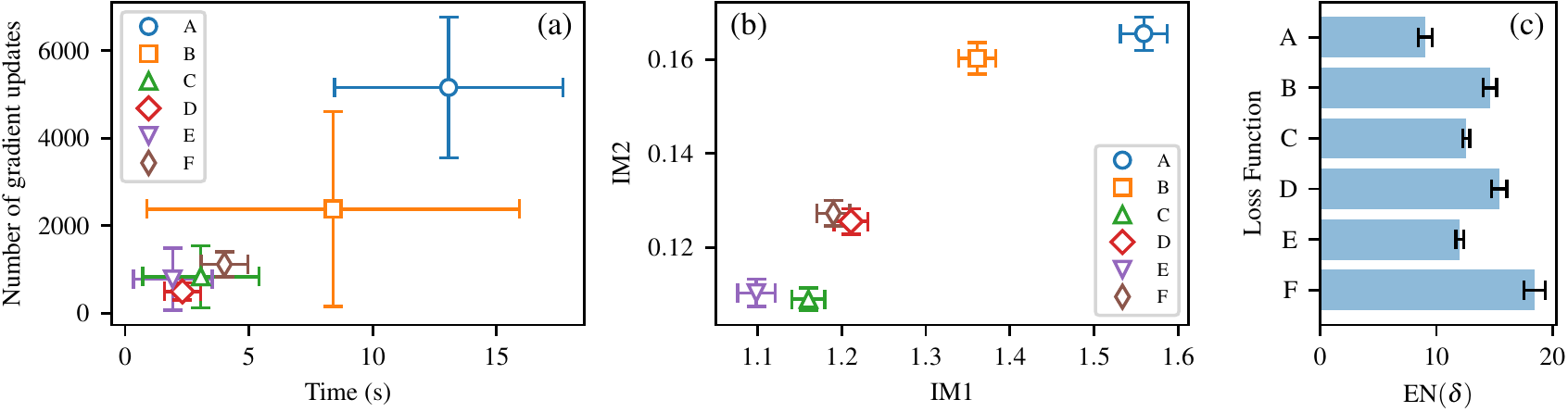}
    \caption{(a) Mean time in seconds and  number of gradient updates needed to find a satisfactory counterfactual for objective functions $A$ to $F$ across all MNIST classes. The error bars represent the standard deviation to illustrate variability between approaches. (b) Mean IM1 and IM2 for objective functions $A$ to $F$ across all MNIST classes (lower is better). The error bars represent the $95$\% confidence bounds. (c) Sparsity measure $\text{EN}(\delta)$ for loss functions $A$ to $F$. The error bars represent the $95$\% confidence bounds.}
    \label{fig:combined}
\end{figure*}

\subsubsection{Speed}
\Cref{fig:combined}(a) shows the mean time and number of gradient steps required to find a satisfactory counterfactual for each objective function. We also show the standard deviations to illustrate the variability between the different loss functions. For loss function $A$, the majority of the time is spent finding a good range for $c$ to find a balance between steering the perturbed instance away from the original class $t_{0}$ and the elastic net regularization. If $c$ is too small, the $L_{1}$ regularization term cancels out the perturbations, but if $c$ is too large, $x_{\text{cf}}$ is not sparse anymore.

The aim of $L_{\text{AE}}$ in loss function $B$ is not to speed up convergence towards a counterfactual instance, but to have $x_{\text{cf}}$ respect the training data distribution. This is backed up by the experiments. The average speed improvement and reduction in the number of gradient updates compared to $A$ of respectively 36\% and 54\% is significant but very inconsistent given the high standard deviation. The addition of $L_{\text{proto}}$ in $C$ however drastically reduces the time and iterations needed by respectively 77\% and 84\% compared to $A$. The combination of $L_{\text{AE}}$ and $L_{\text{proto}}$ in $D$ improves the time to find a counterfactual instance further: $x_{\text{cf}}$ is found 82\% faster compared to $A$, with the number of iterations down by 90\%.
\begin{table}[t]
    \centering
    \caption{Summary statistics with $95$\% confidence bounds for each loss function for the MNIST experiment.}
    \vskip 0.15in
    \resizebox{\columnwidth}{!}{%
    \begin{tabular}{@{}ccccc@{}}
    \toprule
    \textbf{Method} & \textbf{Time (s)} & \textbf{Gradient steps} & \textbf{IM1} & \textbf{IM2 ($\times$10)} \\ \midrule
    A & $13.06 \pm 0.23$ & $5158 \pm 82$ & $1.56 \pm 0.03$ & $1.65 \pm 0.04$ \\
    B & $8.40 \pm 0.38$ & $2380 \pm 113$ & $1.36 \pm 0.02$ & $1.60 \pm 0.03$ \\
    C & $3.06 \pm $0.11 & $835 \pm 36$ & $1.16 \pm 0.02$ & $1.09 \pm 0.02$ \\
    D & $2.31 \pm $0.04 & $497 \pm 10$ & $1.21 \pm 0.02$ & $1.26 \pm 0.03$ \\
    E & $1.93 \pm $0.10 & $777 \pm 44$ & $1.10 \pm 0.02$ & $1.10 \pm 0.03$ \\
    F & $4.01 \pm $0.05 & $1116 \pm 14$ & $1.19 \pm 0.02$ & $1.27 \pm 0.03$ \\ \bottomrule
    \end{tabular}%
    }
    \label{tb:table_mnist}
\end{table}

So far we have assumed access to the model architecture to take advantage of automatic differentiation during the counterfactual search process. $L_{\text{pred}}$ can however form a computational bottleneck for black box models because numerical gradient calculation results in a number of prediction function calls proportionate to the dimensionality of the input features. Consider $A'$ the equivalent of loss function $A$ where we can only query the model's prediction function. $E$ and $F$ remove $L_{\text{pred}}$ which results in approximately a 100x speed up of the counterfactual search process compared to $A'$. The results can be found in the supplementary material.

\subsubsection{Quantitative interpretability}

IM1 peaks for loss function $A$ and improves by respectively 13\% and 26\% as $L_{\text{AE}}$ and $L_{\text{proto}}$ are added (\Cref{fig:combined}(b)). This implies that including $L_{\text{proto}}$ leads to more interpretable counterfactual instances than $L_{\text{AE}}$ which explicitly minimizes the reconstruction error using $\text{AE}$. Removing $L_{\text{pred}}$ in $E$ yields an improvement over $A$ of 29\%. While $L_{\text{pred}}$ encourages the perturbed instance to predict a different class than $t_{0}$, it does not impose any restrictions on the data distribution of $x_{\text{cf}}$. $L_{\text{proto}}$ on the other hand implicitly encourages the perturbed instance to predict $i \neq t_{0}$ while minimizing the distance in latent space to a representative distribution of class $i$.
\begin{figure}[ht]
    \centering
    \includegraphics[width=0.98\columnwidth]{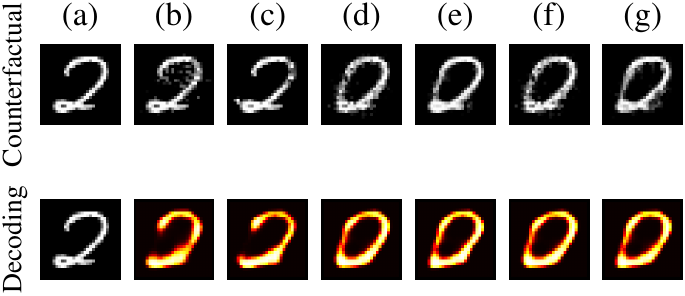}
    \caption{(a) Shows the original instance, (b) to (g) on the first row illustrate counterfactuals generated by using loss functions $A$ to $F$. (b) to (g) on the second row show the reconstructed counterfactuals using $AE$.}
    \label{fig:qual_mnist}
\end{figure}

The picture for IM2 is similar. Adding in $L_{\text{proto}}$ brings IM2 down by 34\% while the combination of $L_{\text{AE}}$ and $L_{\text{proto}}$ only reduces the metric by 24\%. For large values of $K$ the prototypes are further from $\text{ENC}(x_{0})$ resulting in larger initial perturbations towards the counterfactual class. In this case, $L_{\text{AE}}$ ensures the overall distribution is respected which makes the reconstructed images of $\text{AE}_{i}$ and $\text{AE}$ more similar and improves IM2. The impact of $K$ on IM1 and IM2 is illustrated in the supplementary material. The removal of $L_{\text{pred}}$ in $E$ and $F$ has little impact on IM2. This emphasizes that $L_{\text{proto}}$---optionally in combination with $L_{\text{AE}}$---is the dominant term with regards to interpretability.

Finally, performing kernel multiple model comparison tests \cite{lim2019} indicates that counterfactuals generated by methods not including the prototype term ($A$ and $B$) result in high rejection rates for faithfully modelling the predicted class distribution (see supplementary material).

\subsubsection{Visual interpretability}

\Cref{fig:qual_mnist} shows counterfactual examples on the first row and their reconstructions using $\text{AE}$ on the second row for different loss functions. The counterfactuals generated with $A$ or $B$ are sparse but uninterpretable and are still close to the manifold of a $2$. Including $L_{\text{proto}}$ in \Cref{fig:qual_mnist}(d) to (g) leads to a clear, interpretable $0$ which is supported by the reconstructed counterfactuals on the second row. More examples can be found in the supplementary material.

\subsubsection{Sparsity}

The elastic net evaluation metric $\text{EN}(\delta)$ is also the only loss term present in $A$ besides $L_{\text{pred}}$. It is therefore not surprising that $A$ results in the most sparse counterfactuals (\Cref{fig:combined}(c)). The relative importance of sparsity in the objective function goes down as $L_{\text{AE}}$ and $L_{\text{proto}}$ are added. $L_{\text{proto}}$ leads to more sparse counterfactuals than $L_{\text{AE}}$ ($C$ and $E$), but this effect diminishes for large $K$.

\subsection{Breast Cancer Wisconsin (Diagnostic) Dataset}
The second experiment uses the Breast Cancer Wisconsin (Diagnostic) dataset which describes characteristics of cell nuclei in an image and labels them as \textit{malignant} or \textit{benign}. The real-valued features for the nuclei in the image are the mean, error and worst values for characteristics like the radius, texture or area of the nuclei. The dataset contains 569 instances with 30 features each. The first 550 instances are used for training, the last 19 to generate the counterfactuals. For each instance in the test set we generate 5 counterfactuals with different random seeds. Instead of an encoder we use k-d trees to find the prototypes. We evaluate and compare counterfactuals obtained by using the following loss functions:
\begin{equation}\label{eq:bcw_losses}
    \begin{aligned}
        A &= c \cdot L_{\text{pred}} + \beta \cdot L_{1} + L_{2}\\
        B &= c \cdot L_{\text{pred}} + \beta \cdot L_{1} + L_{2} + L_{\text{proto}}\\
        C &= \beta \cdot L_{1} + L_{2} + L_{\text{proto}}
    \end{aligned}
\end{equation}
The model used to classify the instances is a 2 layer feedforward neural network with 40 neurons in each layer. More details can be found in the supplementary material.

\subsubsection{Results}
\Cref{tb:table_bcw} summarizes the findings for the speed and interpretability measures.

\subsubsection{Speed}
$L_{\text{proto}}$ drastically reduces the time and iterations needed to find a satisfactory counterfactual. Loss function $B$ finds $x_{\text{cf}}$ in 13\% of the time needed compared to $A$ while bringing the number of gradient updates down by 91\%. Removing $L_{\text{pred}}$ and solely relying on the prototype to guide $x_{\text{cf}}$ reduces the search time by 92\% and the number of iterations by 93\%.

\subsubsection{Quantitative interpretability}
Including $L_{\text{proto}}$ in the loss function reduces IM1 and IM2 by respectively 55\% and 81\%. Removing $L_{\text{pred}}$ in $C$ results in similar improvements over $A$.

\subsubsection{Sparsity}
Loss function $A$ yields the most sparse counterfactuals. Sparsity and interpretability should however not be considered in isolation. The dataset has 10 attributes (e.g. radius or texture) with 3 values per attribute (mean, error and worst). $B$ and $C$ which include $L_{\text{proto}}$ perturb relatively more values of the same attribute than $A$ which makes intuitive sense. If for instance the worst radius increases, the mean should typically follow as well. The supplementary material supports this statement.

\begin{table}[t]
    \centering
    \caption{Summary statistics with $95$\% confidence bounds for each loss function for the Breast Cancer Wisconsin (Diagnostic) experiment.}
    \vskip 0.15in
    \resizebox{\columnwidth}{!}{%
    \begin{tabular}{@{}ccccc@{}}
    \toprule
    \textbf{Method} & \textbf{Time (s)} & \textbf{Gradient steps} & \textbf{IM1} & \textbf{IM2 ($\times$10)} \\ \midrule
    A & $2.68 \pm 0.20$ & $2752 \pm 203$ & $2.07 \pm 0.16$ & $7.65 \pm 0.79$ \\
    B & $0.35 \pm 0.03$ & $253 \pm 33$ & $0.94 \pm 0.10$ & $1.47 \pm 0.15$ \\
    C & $0.22 \pm 0.02$ & $182 \pm 30$ & $0.88 \pm 0.10$ & $1.41 \pm 0.15$ \\ \bottomrule
    \end{tabular}%
    }
    \label{tb:table_bcw}
\end{table}

\subsection{Adult (Census) Dataset}
\begin{figure}[ht]
    \centering
    \includegraphics[width=0.98\columnwidth]{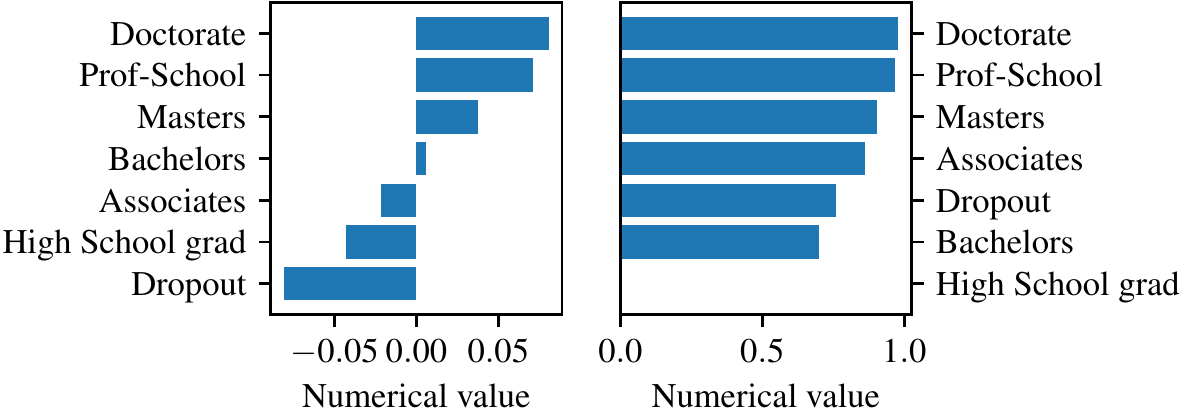}
    \caption{Left: Embedding of the categorical variable ``Education'' in numerical space using association based distance metric (ABDM). Right: Frequency based embedding.}
    \label{fig:f_cat_rank}
\end{figure}

The Adult (Census) dataset consists of individuals described by a mixture of numerical and categorical features. The predictive task is to determine whether a person earns more than \$50k/year. As the dataset contains categorical features, it is important to use a principled approach to define perturbations over these features. \Cref{fig:f_cat_rank} illustrates our approach using the association based distance metric \cite{le2005}(ABDM) to embed the feature ``Education'' into one dimensional numerical space over which perturbations can be defined. The resulting embedding defines a natural ordering of categories in agreement with common sense for this interpretable variable. By contrast, the frequency embedding method as proposed by \cite{dhurandhar2019model} does not capture the underlying relation between categorical values.

Since ABDM infers distances from other variables by computing dissimilarity based on the K-L divergence, it can break down if there is independence between categories. In such cases one can use MVDM \cite{cost1993} which uses the difference between the conditional model prediction probabilities of each category.
A counterfactual example changing categorical features is shown in \Cref{fig:cf_examples}.

\section{Discussion}

In this paper we introduce a model agnostic counterfactual search process guided by class prototypes. We show that including a prototype loss term in the objective results in more interpretable counterfactual instances as measured by two novel interpretability metrics. We demonstrate that prototypes speed up the search process and remove the numerical gradient evaluation bottleneck for black box models thus making our method more appealing for practical applications. By fixing selected features to the original values during the search process we can also obtain \textit{actionable counterfactuals} which describe concrete steps to take to change a model's prediction. To facilitate the practical use of counterfactual explanations we provide an open source library with our implementation of the method \cite{alibi}.

\bibliography{references}

\begin{thebibliography}{29}
\providecommand{\natexlab}[1]{#1}
\providecommand{\url}[1]{\texttt{#1}}
\expandafter\ifx\csname urlstyle\endcsname\relax
  \providecommand{\doi}[1]{doi: #1}\else
  \providecommand{\doi}{doi: \begingroup \urlstyle{rm}\Url}\fi

\bibitem[Banerjee et~al.(2005)Banerjee, Merugu, Dhillon, and
  Ghosh]{Banerjee05clusteringwith}
Banerjee, A., Merugu, S., Dhillon, I.~S., and Ghosh, J.
\newblock Clustering with bregman divergences.
\newblock \emph{Journal of Machine Learning Research}, 6:\penalty0 1705--1749,
  December 2005.
\newblock ISSN 1532-4435.
\newblock URL \url{https://dl.acm.org/citation.cfm?id=1046920.1194902}.

\bibitem[Beck \& Teboulle(2009)Beck and Teboulle]{beck2009}
Beck, A. and Teboulle, M.
\newblock A fast iterative shrinkage-thresholding algorithm for linear inverse
  problems.
\newblock \emph{SIAM Journal on Imaging Sciences}, 2\penalty0 (1):\penalty0
  183--202, March 2009.
\newblock ISSN 1936-4954.
\newblock \doi{10.1137/080716542}.
\newblock URL \url{https://dx.doi.org/10.1137/080716542}.

\bibitem[Bentley(1975)]{Bentley1975}
Bentley, J.~L.
\newblock Multidimensional binary search trees used for associative searching.
\newblock \emph{Communications of the ACM}, 18\penalty0 (9):\penalty0 509--517,
  September 1975.
\newblock ISSN 0001-0782.
\newblock \doi{10.1145/361002.361007}.
\newblock URL \url{https://doi.acm.org/10.1145/361002.361007}.

\bibitem[Bien \& Tibshirani(2011)Bien and Tibshirani]{bien2011}
Bien, J. and Tibshirani, R.
\newblock Prototype selection for interpretable classification.
\newblock \emph{The Annals of Applied Statistics}, 5\penalty0 (4):\penalty0
  2403--2424, 12 2011.
\newblock \doi{10.1214/11-AOAS495}.
\newblock URL \url{https://doi.org/10.1214/11-AOAS495}.

\bibitem[Borg \& Groenen(2005)Borg and Groenen]{borg2005mds}
Borg, I. and Groenen, P.
\newblock \emph{{Modern Multidimensional Scaling: Theory and Applications}}.
\newblock Springer, 2005.

\bibitem[Cost \& Salzberg(1993)Cost and Salzberg]{cost1993}
Cost, S. and Salzberg, S.
\newblock A weighted nearest neighbor algorithm for learning with symbolic
  features.
\newblock \emph{Machine Learning}, 10\penalty0 (1):\penalty0 57--78, Jan 1993.
\newblock ISSN 1573-0565.
\newblock \doi{10.1023/A:1022664626993}.
\newblock URL \url{https://doi.org/10.1023/A:1022664626993}.

\bibitem[Dhurandhar et~al.(2017)Dhurandhar, Iyengar, Luss, and
  Shanmugam]{dhurandhar2017tip}
Dhurandhar, A., Iyengar, V., Luss, R., and Shanmugam, K.
\newblock Tip: Typifying the interpretability of procedures.
\newblock \emph{arXiv preprint arXiv:1706.02952}, 2017.
\newblock URL \url{https://arxiv.org/abs/1706.02952}.

\bibitem[Dhurandhar et~al.(2018)Dhurandhar, Chen, Luss, Tu, Ting, Shanmugam,
  and Das]{dhurandhar2018}
Dhurandhar, A., Chen, P.-Y., Luss, R., Tu, C.-C., Ting, P., Shanmugam, K., and
  Das, P.
\newblock Explanations based on the missing: Towards contrastive explanations
  with pertinent negatives.
\newblock In \emph{Advances in Neural Information Processing Systems 31}, pp.\
  592--603. 2018.
\newblock URL
  \url{https://papers.nips.cc/paper/7340-explanations-based-on-the-missing-towards-contrastive-explanations-with-pertinent-negatives}.

\bibitem[Dhurandhar et~al.(2019)Dhurandhar, Pedapati, Balakrishnan, Chen,
  Shanmugam, and Puri]{dhurandhar2019model}
Dhurandhar, A., Pedapati, T., Balakrishnan, A., Chen, P.-Y., Shanmugam, K., and
  Puri, R.
\newblock Model agnostic contrastive explanations for structured data.
\newblock \emph{arXiv preprint arXiv:1906.00117}, 2019.
\newblock URL \url{https://arxiv.org/abs/1906.00117}.

\bibitem[Dua \& Graff(2017)Dua and Graff]{Dua2019}
Dua, D. and Graff, C.
\newblock {UCI} machine learning repository, 2017.
\newblock URL \url{https://archive.ics.uci.edu/ml}.

\bibitem[Gurumoorthy et~al.(2017)Gurumoorthy, Dhurandhar, and
  Cecchi]{gurumoorthy2017protodash}
Gurumoorthy, K.~S., Dhurandhar, A., and Cecchi, G.
\newblock Protodash: fast interpretable prototype selection.
\newblock \emph{arXiv preprint arXiv:1707.01212}, 2017.
\newblock URL \url{https://arxiv.org/abs/1707.01212}.

\bibitem[Jiang et~al.(2018)Jiang, Kim, Guan, and Gupta]{jiang2018}
Jiang, H., Kim, B., Guan, M., and Gupta, M.
\newblock To trust or not to trust a classifier.
\newblock In \emph{Advances in Neural Information Processing Systems 31}, pp.\
  5541--5552. 2018.
\newblock URL
  \url{https://papers.nips.cc/paper/7798-to-trust-or-not-to-trust-a-classifier}.

\bibitem[Kaufmann \& Rousseeuw(1987)Kaufmann and
  Rousseeuw]{kaufmanl1987clustering}
Kaufmann, L. and Rousseeuw, P.
\newblock Clustering by means of medoids.
\newblock \emph{Data Analysis based on the L1-Norm and Related Methods}, pp.\
  405--416, 01 1987.

\bibitem[Kim et~al.(2016)Kim, Khanna, and Koyejo]{beenkim2016criticism}
Kim, B., Khanna, R., and Koyejo, O.~O.
\newblock Examples are not enough, learn to criticize! criticism for
  interpretability.
\newblock In \emph{Advances in Neural Information Processing Systems 29}, pp.\
  2280--2288. 2016.
\newblock URL
  \url{https://papers.nips.cc/paper/6300-examples-are-not-enough-learn-to-criticize-criticism-for-interpretability}.

\bibitem[Klaise et~al.()Klaise, Van~Looveren, Vacanti, and Coca]{alibi}
Klaise, J., Van~Looveren, A., Vacanti, G., and Coca, A.
\newblock Alibi: Algorithms for monitoring and explaining machine learning
  models.
\newblock URL \url{https://github.com/SeldonIO/alibi}.

\bibitem[Laugel et~al.(2018)Laugel, Lesot, Marsala, Renard, and
  Detyniecki]{thibault2018}
Laugel, T., Lesot, M.-J., Marsala, C., Renard, X., and Detyniecki, M.
\newblock Comparison-based inverse classification for interpretability in
  machine learning.
\newblock In \emph{Information Processing and Management of Uncertainty in
  Knowledge-Based Systems. Theory and Foundations}, pp.\  100--111. Springer
  International Publishing, 2018.
\newblock ISBN 978-3-319-91473-2.
\newblock URL \url{https://hal.sorbonne-universite.fr/hal-01905982}.

\bibitem[Le \& Ho(2005)Le and Ho]{le2005}
Le, S.~Q. and Ho, T.~B.
\newblock An association-based dissimilarity measure for categorical data.
\newblock \emph{Pattern Recognition Letters}, 26\penalty0 (16):\penalty0 2549
  -- 2557, 2005.
\newblock ISSN 0167-8655.
\newblock \doi{https://doi.org/10.1016/j.patrec.2005.06.002}.
\newblock URL
  \url{http://www.sciencedirect.com/science/article/pii/S0167865505001686}.

\bibitem[LeCun \& Cortes(2010)LeCun and Cortes]{lecun2010mnist}
LeCun, Y. and Cortes, C.
\newblock {MNIST} handwritten digit database.
\newblock 2010.
\newblock URL \url{http://yann.lecun.com/exdb/mnist/}.

\bibitem[Lim et~al.(2019)Lim, Yamada, Sch\"{o}lkopf, and Jitkrittum]{lim2019}
Lim, J.~N., Yamada, M., Sch\"{o}lkopf, B., and Jitkrittum, W.
\newblock Kernel stein tests for multiple model comparison.
\newblock In \emph{Advances in Neural Information Processing Systems 32}, pp.\
  2240--2250. Curran Associates, Inc., 2019.
\newblock URL
  \url{http://papers.nips.cc/paper/8496-kernel-stein-tests-for-multiple-model-comparison.pdf}.

\bibitem[Lundberg \& Lee(2017)Lundberg and Lee]{lundberg2017}
Lundberg, S.~M. and Lee, S.-I.
\newblock A unified approach to interpreting model predictions.
\newblock In \emph{Advances in Neural Information Processing Systems 30}, pp.\
  4765--4774. 2017.
\newblock URL
  \url{https://papers.nips.cc/paper/7062-a-unified-approach-to-interpreting-model-predictions}.

\bibitem[Luss et~al.(2019)Luss, Chen, Dhurandhar, Sattigeri, Shanmugam, and
  Tu]{luss2019generating}
Luss, R., Chen, P.-Y., Dhurandhar, A., Sattigeri, P., Shanmugam, K., and Tu,
  C.-C.
\newblock Generating contrastive explanations with monotonic attribute
  functions.
\newblock \emph{arXiv preprint arXiv:1905.12698}, 2019.
\newblock URL \url{https://arxiv.org/abs/1905.12698}.

\bibitem[Molnar(2019)]{molnar2019}
Molnar, C.
\newblock \emph{Interpretable Machine Learning}.
\newblock 2019.
\newblock \url{https://christophm.github.io/interpretable-ml-book/}; accessed
  22-January-2020.

\bibitem[Mothilal et~al.(2020)Mothilal, Sharma, and Tan]{Mothilal_2020}
Mothilal, R.~K., Sharma, A., and Tan, C.
\newblock Explaining machine learning classifiers through diverse
  counterfactual explanations.
\newblock \emph{Proceedings of the 2020 Conference on Fairness, Accountability,
  and Transparency}, Jan 2020.
\newblock \doi{10.1145/3351095.3372850}.
\newblock URL \url{http://dx.doi.org/10.1145/3351095.3372850}.

\bibitem[Ribeiro et~al.(2016)Ribeiro, Singh, and Guestrin]{ribeiro2016}
Ribeiro, M.~T., Singh, S., and Guestrin, C.
\newblock ``{W}hy should {I} trust you'': Explaining the predictions of any
  classifier.
\newblock In \emph{Proceedings of the 22Nd ACM SIGKDD International Conference
  on Knowledge Discovery and Data Mining}, pp.\  1135--1144, 2016.
\newblock ISBN 978-1-4503-4232-2.
\newblock \doi{10.1145/2939672.2939778}.
\newblock URL \url{https://doi.acm.org/10.1145/2939672.2939778}.

\bibitem[Rumelhart et~al.(1986)Rumelhart, Hinton, and Williams]{rumelhart1986}
Rumelhart, D.~E., Hinton, G.~E., and Williams, R.~J.
\newblock Learning internal representations by error propagation.
\newblock In \emph{Parallel Distributed Processing: Explorations in the
  Microstructure of Cognition, {V}ol. 1}, pp.\  318--362. MIT Press, Cambridge,
  MA, USA, 1986.
\newblock ISBN 0-262-68053-X.
\newblock URL \url{https://dl.acm.org/citation.cfm?id=104279.104293}.

\bibitem[Snell et~al.(2017)Snell, Swersky, and Zemel]{snell2017}
Snell, J., Swersky, K., and Zemel, R.
\newblock Prototypical networks for few-shot learning.
\newblock In \emph{Advances in Neural Information Processing Systems 30}, pp.\
  4077--4087. 2017.
\newblock URL
  \url{https://papers.nips.cc/paper/6996-prototypical-networks-for-few-shot-learning}.

\bibitem[Takigawa et~al.(2009)Takigawa, Kudo, and Nakamura]{TAKIGAWA2009101}
Takigawa, I., Kudo, M., and Nakamura, A.
\newblock Convex sets as prototypes for classifying patterns.
\newblock \emph{Engineering Applications of Artificial Intelligence},
  22\penalty0 (1):\penalty0 101 -- 108, 2009.
\newblock ISSN 0952-1976.
\newblock \doi{https://doi.org/10.1016/j.engappai.2008.05.012}.
\newblock URL
  \url{https://www.sciencedirect.com/science/article/pii/S0952197608001589}.

\bibitem[Wachter et~al.(2018)Wachter, Mittelstadt, and Russell]{wachter2018}
Wachter, S., Mittelstadt, B., and Russell, C.
\newblock Counterfactual explanations without opening the black box: Automated
  decisions and the {GDPR}.
\newblock \emph{Harvard journal of law \& technology}, 31:\penalty0 841--887,
  04 2018.
\newblock URL \url{https://arxiv.org/abs/1711.00399}.

\bibitem[Zou \& Hastie(2005)Zou and Hastie]{zou2005en}
Zou, H. and Hastie, T.
\newblock Regularization and variable selection via the elastic net.
\newblock \emph{Journal of the Royal Statistical Society. Series B (Statistical
  Methodology)}, 67\penalty0 (2):\penalty0 301--320, 2005.
\newblock ISSN 13697412, 14679868.
\newblock URL \url{https://www.jstor.org/stable/3647580}.

\end{thebibliography}
\bibliographystyle{icml2020}

\appendix
\section{Breast Cancer Wisconsin experiment details and results}

The classification model used to classify the cell nuclei into the \textit{malignant} or \textit{benign} categories is a 2 layer feedforward neural network with 40 neurons and $\text{ReLU}$ activations in each layer. The model is trained on standardized features with stochastic gradient descent for 500 epochs with batch size 128 and reaches 100\% accuracy on the test set.

The class specific autoencoders used to evaluate IM1 and IM2 consist of 3 dense layers in the encoder with respectively 20, 10 and 6 neurons for each layer. The first 2 layers have $\text{ReLU}$ activations whilst the last one has a linear activation. The dense layers in the decoder contain 10 and 20 neurons followed by a linear layer projecting the reconstructed instance back to the input feature space. The autoencoders are optimized with Adam and trained for 500 epochs on batches of 128 instances with the mean squared error between the original and reconstructed instance as the loss function.

\begin{figure*}[ht]
    \centering
    \includegraphics[width=0.95\textwidth]{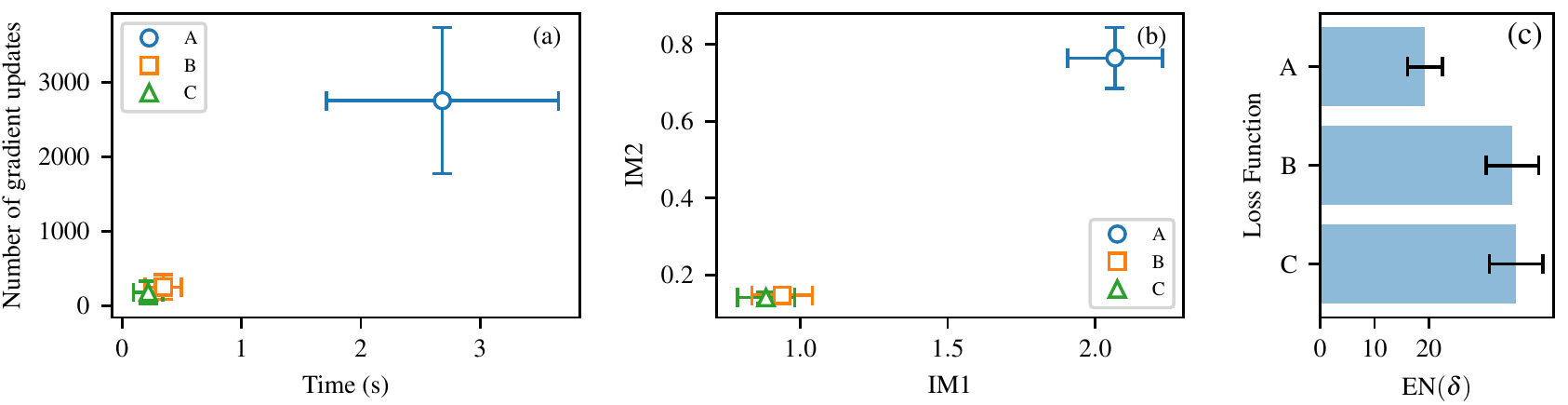}
    \caption{(a) Mean time in seconds and  number of gradient updates needed to find a satisfactory counterfactual for objective functions $A$, $B$ and $C$ for the Breast Cancer Wisconsin dataset. The error bars represent the standard deviation to illustrate variability between approaches. (b) Mean IM1 and IM2 for objective functions $A$, $B$ and $C$ (lower is better). The error bars represent the $95$\% confidence bounds. (c) Sparsity measure $\text{EN}(\delta)$ for loss functions $A$, $B$ and $C$. The error bars represent the $95$\% confidence bounds.}
    \label{fig:combined_cancer}
\end{figure*}

Similar to the MNIST experiment, parameters $c$, $\kappa$ and $\beta$ are kept constant throughout the experiments at 1, 0 and 0.1. The results of the experiment are visualized in \Cref{fig:combined_cancer}.

\begin{figure*}[ht]
    \centering
    \includegraphics[width=0.95\textwidth]{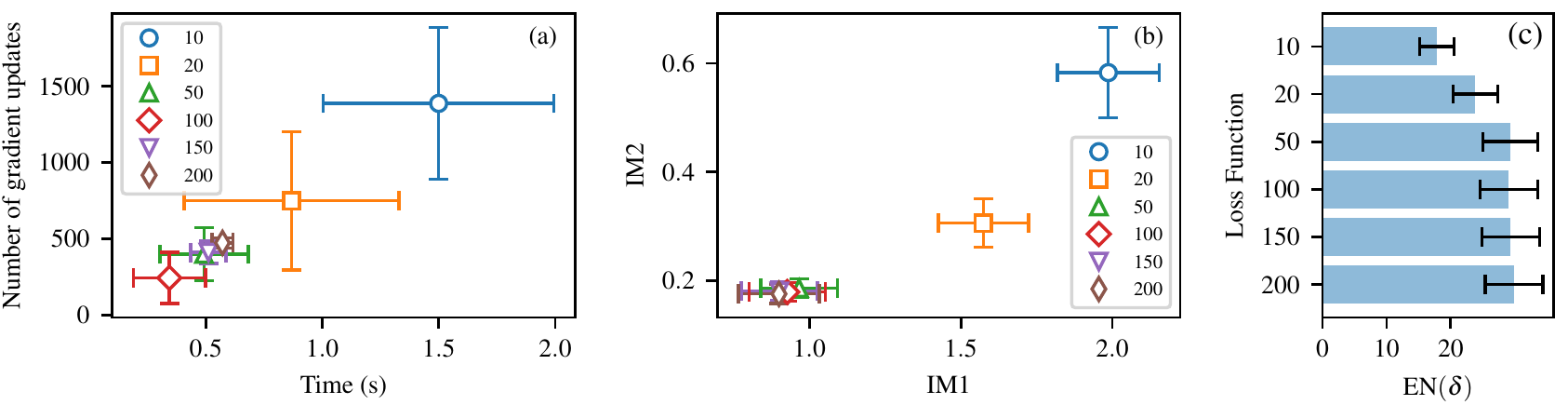}
    \caption{Impact of $\theta$. (a) Mean time in seconds and  number of gradient updates needed to find a satisfactory counterfactual for objective function $B$ with different values of $\theta$ (10, 20, 50, 100, 150, 200) for the Breast Cancer Wisconsin dataset. The error bars represent the standard deviation to illustrate variability between approaches. (b) Mean IM1 and IM2 for objective function $B$ for different values of $\theta$ (lower is better). The error bars represent the $95$\% confidence bounds. (c) Sparsity measure $\text{EN}(\delta)$ for loss functions $B$ and different $\theta$ values. The error bars represent the $95$\% confidence bounds.}
    \label{fig:combined_cancer_theta}
\end{figure*}

\begin{figure*}[!ht]
    \centering
    \includegraphics[width=0.95\textwidth]{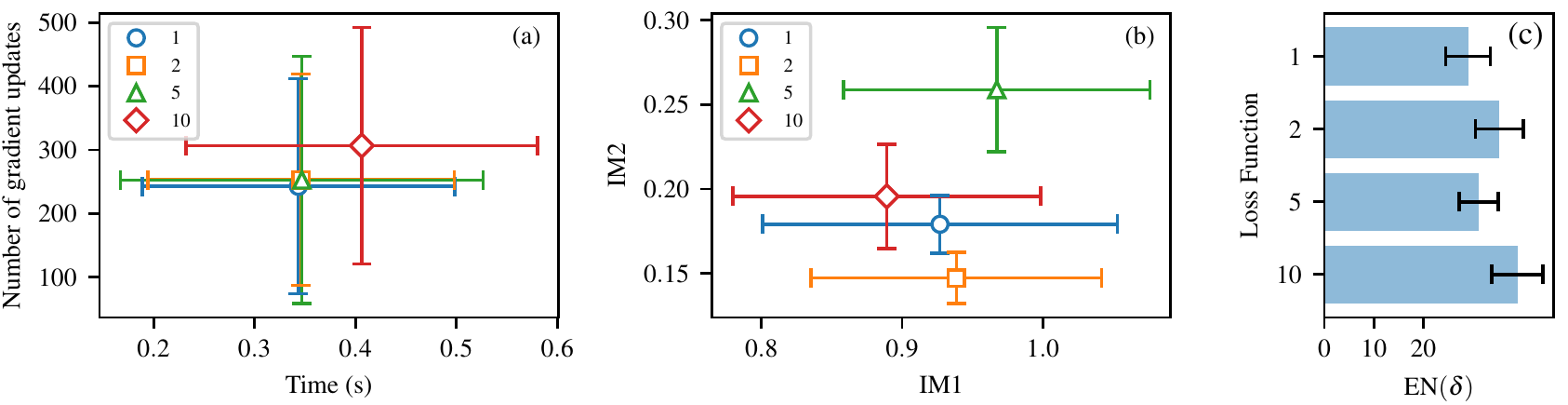}
    \caption{Impact of $k$. (a) Mean time in seconds and  number of gradient updates needed to find a satisfactory counterfactual for objective function $B$ with different values for the $k$th nearest instance in each class ($k$ set to 1, 2, 5 and 10) which is used to define the prototype for the Breast Cancer Wisconsin dataset. The error bars represent the standard deviation to illustrate variability between approaches. (b) Mean IM1 and IM2 for objective function $B$ for different values of $k$ (lower is better). The error bars represent the $95$\% confidence bounds. (c) Sparsity measure $\text{EN}(\delta)$ for loss function $B$ and different $k$ values. The error bars represent the $95$\% confidence bounds.}
    \label{fig:combined_cancer_k}
\end{figure*}

We also study the impact of different values for hyperparameters $\theta$ and $k$ which is visualized in \Cref{fig:combined_cancer_theta,fig:combined_cancer_k}. The figures show that a broad range of values for both $\theta$ and $k$ work well.

\begin{figure}[ht]
    \centering
    \includegraphics[width=\columnwidth]{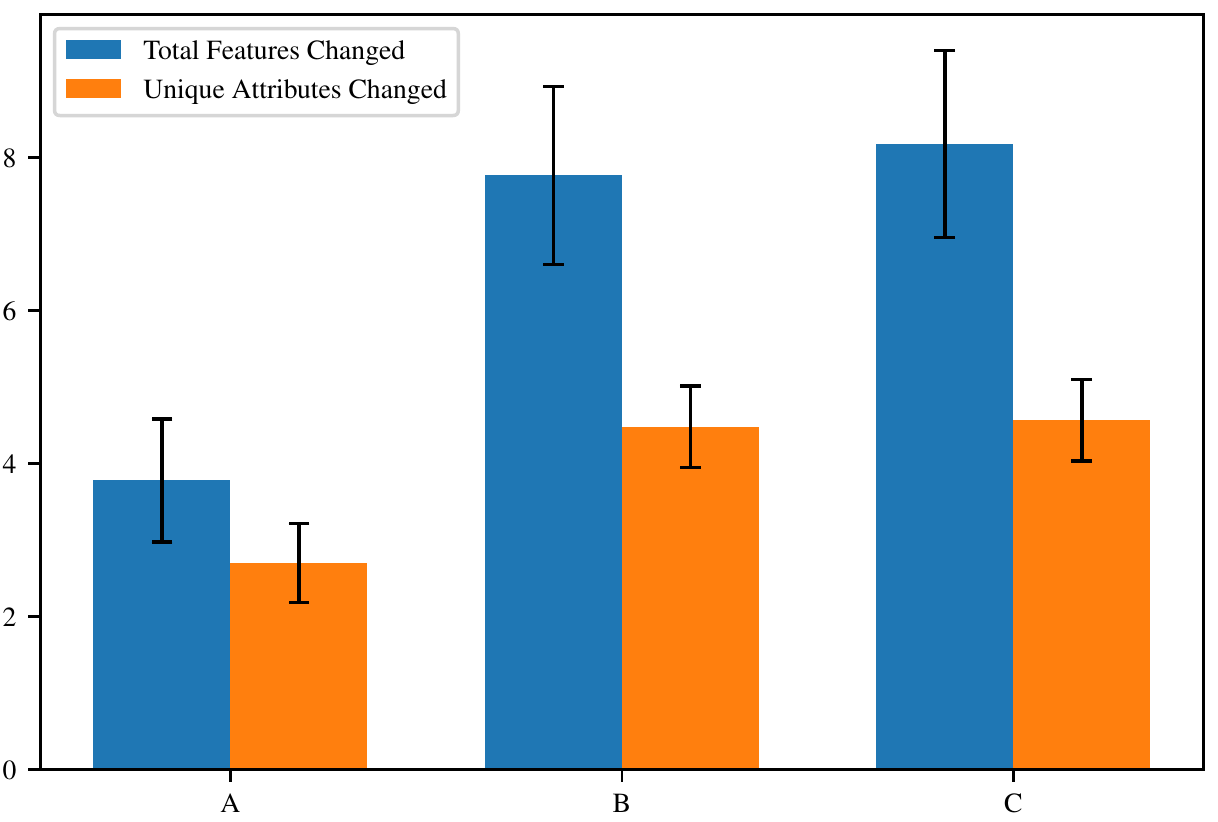}
    \caption{Total number of features and unique number of attributes changed by more than 1 standard deviation in $x_{\text{cf}}$ compared to $x_{0}$ for loss functions $A$, $B$ and $C$. The error bars represent the $95$\% confidence bound. $A$ leads to sparser counterfactuals than $B$ and $C$ but perturbs relatively more unique attributes (e.g. radius or texture) while $B$ and $C$ perturb relatively more features of the same attribute (e.g. mean or worst value of the attribute).}
    \label{fig10}
\end{figure}

Finally we study the number of unique features changed by each method. The results are visualized in \Cref{fig10}.

All experiments were run on a Thinkpad T480 with an  Intel Core i7-8550U Processor.

\section{MNIST experiment details}\label{sec:mnist}

The classification model consists of 2 convolutional layers with respectively 64 and 32 $2\times 2$ filters and $\text{ReLU}$ activations. Each convolutional layer is followed by a $2\times 2$ max-pooling layer. Dropout with fraction 30\% is applied during training. The output of the second pooling layer is flattened and fed into a fully connected layer of size 256 with $\text{ReLU}$ activation and 50\% dropout. This dense layer is followed by a softmax output layer over the 10 classes. The model is trained with an Adam optimizer for 3 epochs with batch size 64 on MNIST images scaled to $[-0.5, 0.5]$ and reaches a test accuracy of 98.6\%. 

The autoencoder used in objective functions $B$ to $F$ has 3 convolutional layers in the encoder. The first 2 contain 16 $3\times 3$ filters and $\text{ReLU}$ activations and are followed by a $2\times 2$ max-pooling layer which feeds into a convolution layer with 1 $3\times 3$ filter and linear activation. The decoder takes the encoded instance as input and feeds it into a convolutional layer with 16 $3\times 3$ filters and $\text{ReLU}$ activations, followed by a $2\times 2$ upsampling layer and again the same convolutional layer. The final convolutional is similar to the last layer in the encoder. All the convolutions have \textit{same} padding. The autoencoder is trained with an Adam optimizer for 4 epochs with batch size 128 and uses the mean squared error between the original and reconstructed instance as the loss function.

The class specific autoencoders used to evaluate IM1 and IM2 consist of 3 convolutional layers with $3\times 3$ filters and $\text{ReLU}$ activations in the encoder, each followed by $2\times 2$ max-pooling layers. The first one contains 16 filters while the others have 8 filters. The decoder follows the same architecture in reversed order and with upsampling instead of max-pooling. The autoencoder is trained with an Adam optimizer for 30 epochs and batch size 128.

Parameters $c$, $\kappa$, $\beta$ and $\gamma$ are kept constant throughout the experiments at 1, 0, 0.1 and 100. Both $L_{\text{AE}}$ and $L_{\text{proto}}$ are reconstruction errors, but $L_{\text{AE}}$ works on the full input feature space while $L_{\text{proto}}$ operates on the compressed latent space. $\theta$ is therefore set at 200 for loss functions $C$ and $E$, and 100 if used in combination with $L_{\text{AE}}$ in $D$ and $F$.

All experiments were run on a Thinkpad T480 with an Intel Core i7-8550U Processor.

\section{MNIST additional results}
\Cref{tb:table_sm} shows the impact on the speed of counterfactual search if the model is only exposed as a black box requiring the calculation of numerical gradients.

\begin{table}[ht]
    \centering
    \caption{Mean time in seconds needed to compute 100 optimization steps for objective functions $A'$, $E$ and $F$ with $95$\% confidence bounds. $A'$ is the equivalent of $A$ without access to the model architecture. As a result, we can only query the prediction function and need to evaluate gradients numerically. One test instance is used for each class in MNIST.}
    \vskip 0.15in
    \resizebox{0.4\columnwidth}{!}{%
    \begin{tabular}{@{}ccccc@{}}
    \toprule
    \textbf{Method} & \textbf{Time (s)} \\ \midrule
    A' & $54.64 \pm 1.28$ \\
    E & $0.53 \pm 0.04$ \\
    F & $0.72 \pm 0.01$ \\ \bottomrule
    \end{tabular}%
    }
    \label{tb:table_sm}
\end{table}

\begin{figure*}[!htb]
    \centering
    \includegraphics[width=0.95\textwidth]{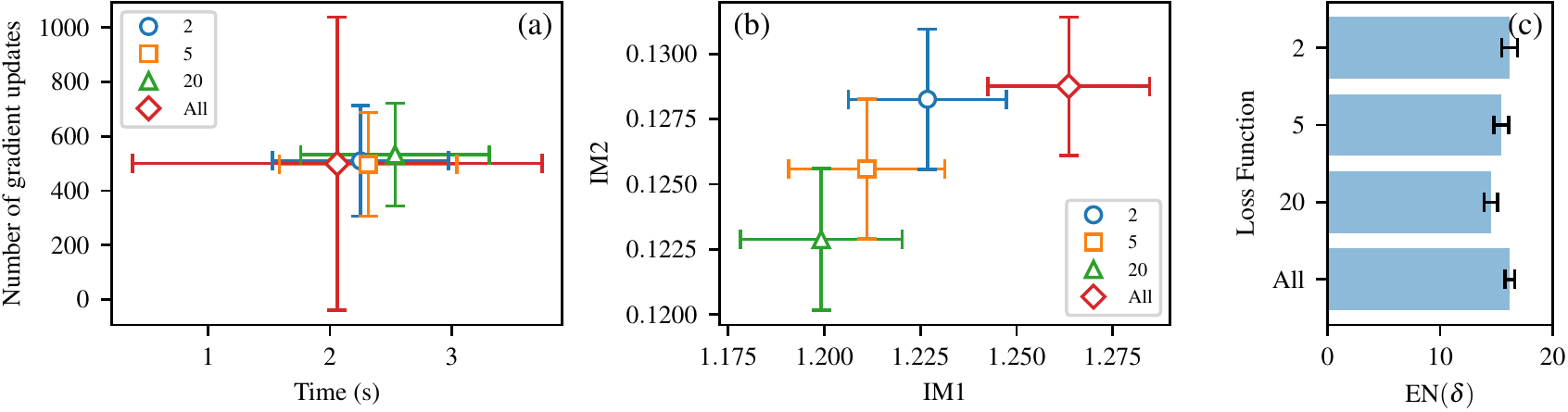}
    \caption{Impact of $K$. (a) Mean time in seconds and  number of gradient updates needed to find a satisfactory counterfactual for objective function $D$ with different values for the number of nearest encoded instances $K$ used for the prototype of each class ($K$ set to 2, 5, 20 and all instances belonging to a class) for the MNIST dataset. The error bars represent the standard deviation to illustrate variability between approaches. (b) Mean IM1 and IM2 for objective function $D$ for different values of $K$ (lower is better). The error bars represent the $95$\% confidence bounds. The interpretability of the counterfactual instances initially improves with increasing $K$ as the prototype includes more instances away from the decision boundary between the original and counterfactual class, more closely representing a typical encoded instance of the counterfactual class. If on the other hand all instances of a class are included in the prototype then this prototype is also defined by instances close to other classes other than the original class and the interpretability worsens. (c) Sparsity measure $\text{EN}(\delta)$ for loss function $D$ and different $K$ values. The error bars represent the $95$\% confidence bounds.}
    \label{fig:combined_k}
\end{figure*}

\Cref{fig:combined_k} shows the effect of the parameter $K$ on the speed, quality and sparsity of counterfactuals generated on the MNIST dataset.

\Cref{fig:mnist_examples,fig:mnist_examples_cont} show additional examples of counterfactual instances on the MNIST dataset generated with various objective functions.

\section{Categorical feature embedding details}
Our goal is to embed every categorical feature into one-dimensional numerical space in which perturbations can be defined and thus counterfactual search performed. We want the embedding to reflect the underlying relation between categorical values.

We proceed in two steps. First, for each categorical variable with $k$ values we want to find a $k\times k$ matrix describing distances between the values. Second, we use multidimensional scaling to transform the pairwise distances into an embedding of feature values into one-dimensional space. We consider the Modified Value Distance Metric (MVDM) and the Association Based Distance Metric (ABDM) to find the required pairwise distance matrix for each categorical variable.

\subsection{Pairwise distances}
MVDM \cite{cost1993} takes as input the data $X$, the labels or predictions of a classifier $y$ and a parameter $\alpha$. In the following we assume that each column of $X$ corresponds either to a single numerical variable or a single categorical variable with ordinal encoding (one-hot encoded categorical variables can be trivially converted to ordinal encoding if necessary). Let $n$ be the number of classes in $y$ and let $v_1,v_2$ be two categories of a categorical feature $v$. Then the distance between the two values is defined as
\begin{equation}
    d(v_1, v_2) = \sum_{i=1}^{n}\left\vert \frac{c_1^i}{c_1} - \frac{c_2^i}{c_2}\right\vert^\alpha,
\end{equation}
where $c_1^i$ is the number of instances with the value $v_1$ that were classified into class $i$ and $c_1$ the total number of instances with the value $v_1$ (likewise for $c_2^i$ and $c_2$). We set $\alpha=1$. The interpretation of this metric is that we want two categories to be more similar if they appear with the same relative frequency across all prediction classes.

ABDM \cite{le2005} takes as input the data $X$. Let $x_1,\dots,x_m$ denote each categorical feature in $X$. The distance between two values of a feature $x_i$ is defined as
\begin{equation}
    d(v_1, v_2) = \sum_{j\neq i}\Psi\left(p(x_j\vert x_i=v_1),p(x_j\vert x_i=v_2))\right),
\end{equation}
where $\Psi$ is a dissimilarity function between two probability distributions and $p(x_j\vert x_i)$ is the conditional probability distribution of feature $x_j$ given feature $x_i$. Thus the distance between two values $v_1,v_2$ of a categorical feature is directly proportional to the distances between the conditional probability distributions of other features given $v_1,v_2$. Following \citet{le2005} we use the Kullback-Leibler divergence as the dissimilarity function $\Psi$. In practice, we also discretize every numerical feature in $X$, i.e. we calculate a histogram and map each original value to a bin. This allows the method to use both categorical and numerical features for inferring categorical distances.

\subsection{Multidimensional scaling}
After inferring the pairwise distances between categories we use multidimensional scaling to embed each category into 2-dimensional Euclidean space. We then use the norms of the embeddings as the one-dimensional numerical values of categories. We use the datapoint with the largest Frobenius norm in the embedded space as the origin. Finally we scale the numerical values for each category using either standard or min-max scaling to ensure the embedded categorical features are in the same range as pre-processed numerical features.

\section{Multiple model comparison test description and results on MNIST}

\begin{table*}[!t]
    \centering
    \caption{Total number of counterfactual MNIST instances generated for each loss function and per counterfactual predicted class. We highlight the lowest number of instances per class which defines the sample size for performing the per-class test.}
    \vskip 0.15in
    \resizebox{0.6\textwidth}{!}{%
    \begin{tabular}{@{}cccccccccccc@{}}
    \toprule
    \textbf{Method} & \textbf{Total} & \textbf{0} & \textbf{1} & \textbf{2} & \textbf{3} & \textbf{4} & \textbf{5} & \textbf{6} & \textbf{7} & \textbf{8} & \textbf{ 9}\\ 
    \midrule
    A & 5375 & 81 & 46 & 737 & 884 & 207 & 610 & 373 & 344 & 1398 & 695 \\
    B & 5377 & \textbf{48} & \textbf{8} & 505 & 708 & \textbf{98} & \textbf{155} & \textbf{197} & \textbf{233} & 2819 & \textbf{606} \\
    C & 5206 & 398 & 88 & 231 & 788 & 397 & 663 & 386 & 632 & 606 & 1017 \\
    D & 4425 & 247 & 54 & \textbf{201} & \textbf{681} & 286 & 529 & 352 & 515 & \textbf{600} & 960 \\
    E & 5232 & 399 & 83 & 231 & 788 & 394 & 664 & 387 & 659 & 609 & 1018 \\
    F & 5120 & 379 & 90 & 226 & 762 & 390 & 627 & 392 & 633 & 604 & 1017 \\
    \bottomrule
    \end{tabular}%
    }
    \label{tb:test_1}
\end{table*}

We want to evaluate the generated counterfactual instances from each loss function on how well they model the counterfactual predicted class distribution. To do this we perform a kernel multiple model comparison test \cite{lim2019} using the open source library from the authors.\footnote{\url{https://github.com/wittawatj/model-comparison-test}}

The method compares $l>2$ models (for us $l=6$ corresponding to loss functions $A$ to $F$) on their relative fit to the data generating distribution (the test set of MNIST). The goal is to decide whether each candidate model is worse than the best one in the candidate list. Thus the test proceeds in two steps. In step one, a reference model is selected such that it is the model minimizing the discrepancy measure (we use maximum mean discrepancy or MMD) between samples generated from it and the data generating distribution (note that this is a random variable). In step two the actual test is performed with the null hypothesis that each model not selected as the reference model in step one is worse than the reference model given that the reference model was selected. Thus the hypothesis is conditional on the selection event of the reference model.

We run the tests on two flavours of the method presented in \citet{lim2019}. RelMulti partitions the sample into two disjoint sets so that step one and two for choosing the reference model and performing the test are done on independent sets. RelPSI uses the same sample for both steps. RelPSI is presented as an alternative for controlling the false positive rate which is lost during the RelMulti procedure (when the selection step is wrong, the test will give a lower true positive rate if using RelMulti).

To generate samples, for each loss function $A$ to $F$ we generate a counterfactual instance from real instances from the test set of MNIST and record the original predicted class and the counterfactual predicted class. The total number of samples generated per counterfactual prediction class for each loss is summarized in \Cref{tb:test_1}. The original instances used were the same for each loss function, the discrepancy between the total numbers is due to constraints on compute.

Note that because we can't control which class the counterfactual instance will end up in some numbers are very low (e.g. it is very hard for any digit to become a counterfactual $1$). This has an impact on the sample size of the per-class tests, e.g. for the counterfactual class $1$ loss function $B$ only returned $8$ instances classified as $1$ by the model which defines the sample size for the test of class $1$ across all loss functions.

For a single test for each class we thus set the sample size to be the minimum number of instances found across all losses (highlighted in \Cref{tb:test_1}). For each loss that returned a higher number of instances we sample the same number without replacement. For the comparison with respect to real instances we sample the same number without replacement from the MNIST test set for that class.

Following \citet{lim2019} we use the Inverse Multiquadratic kernel on 256 features extracted before the softmax layer of the CNN described in \Cref{sec:mnist}. For RelMulti the sample is split in proportion 50:50 to perform the fitting and the test respectively.

We perform the tests 100 times on each class and following \citet{lim2019} measure the selection and rejection rates of each loss function to gauge the quality of generated counterfactuals.

\begin{figure*}[ht]
    \centering
    \includegraphics[width=0.7\textwidth]{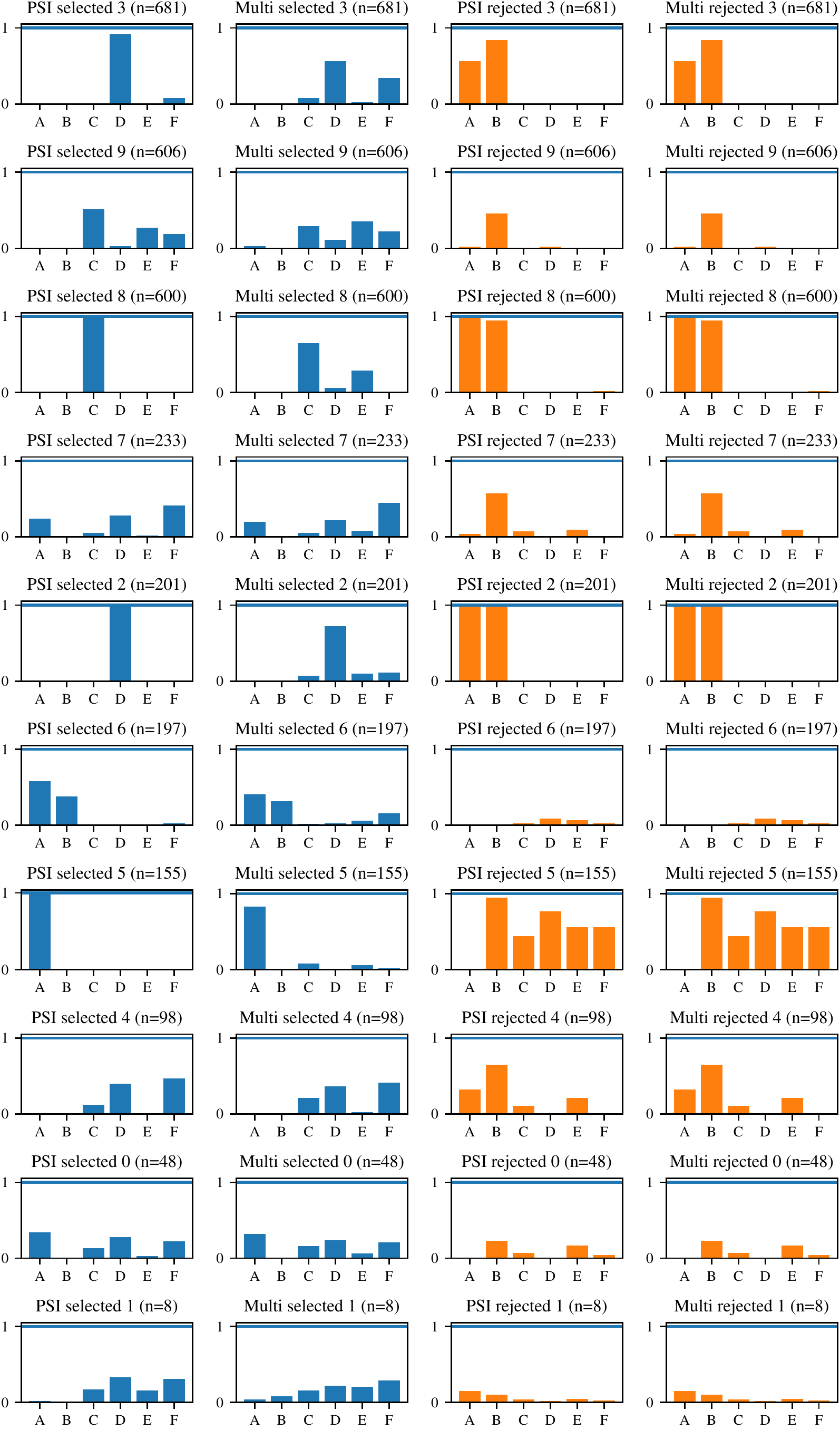}
    \caption{Multiple model comparison test results. Each row shows results for each class of counterfactual instances in decreasing order of available sample size. We report the selection and rejection rate for two methods---RelMulti and RelPSI.}
    \label{fig:sm_testres}
\end{figure*}

\begin{figure*}[ht]
    \centering
    \includegraphics[width=0.47\textwidth]{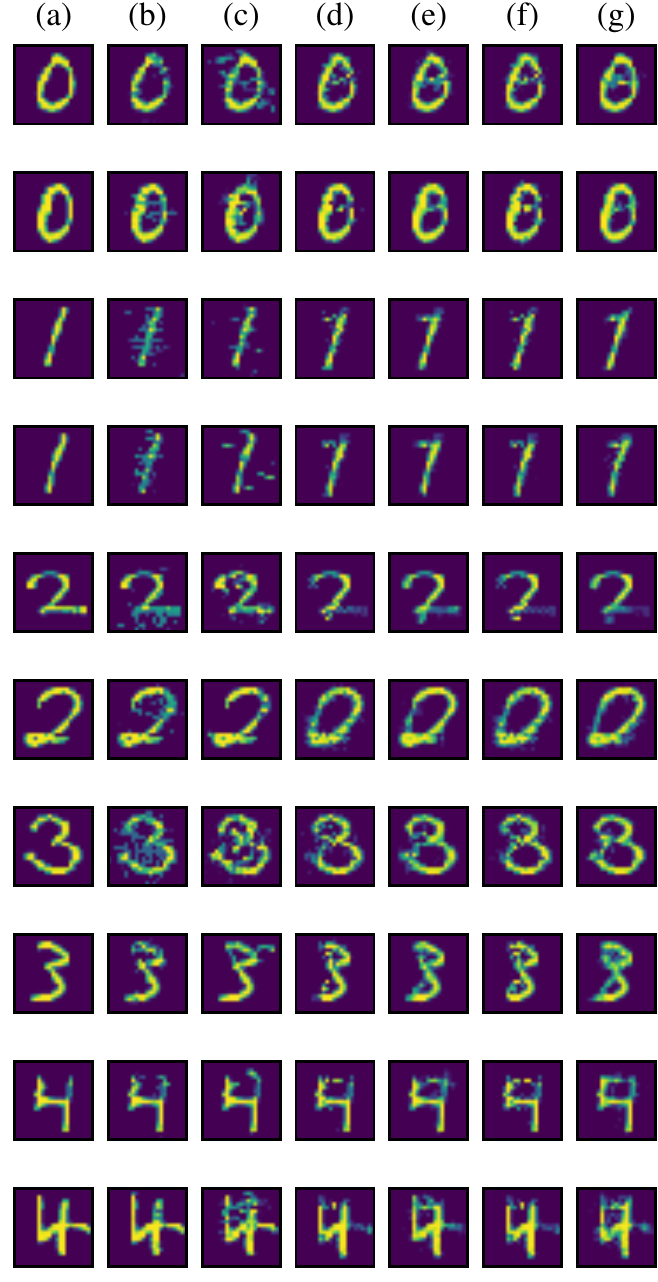}
    \caption{(a) Shows the original instance, (b) to (g) illustrate counterfactuals generated by using loss functions $A$, $B$, $C$, $D$, $E$ and $F$.}
    \label{fig:mnist_examples}
\end{figure*}

\begin{figure*}[ht]
    \centering
    \includegraphics[width=0.47\textwidth]{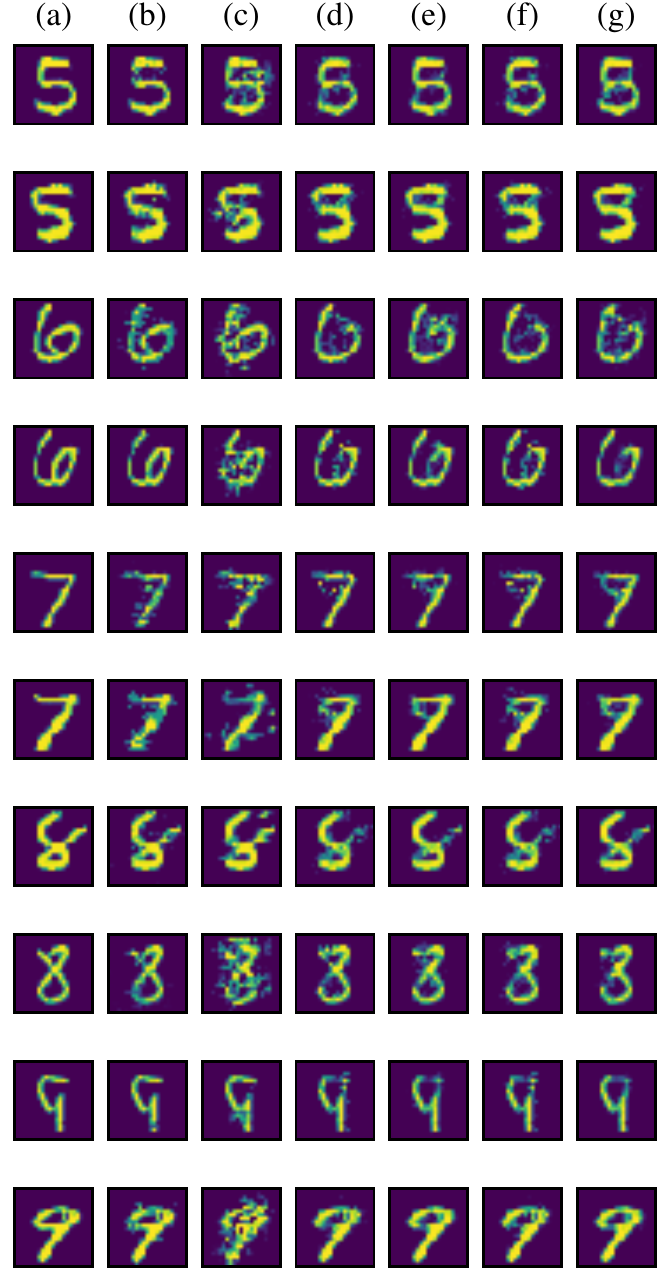}
    \caption{(a) Shows the original instance, (b) to (g) illustrate counterfactuals generated by using loss functions $A$, $B$, $C$, $D$, $E$ and $F$.}
    \label{fig:mnist_examples_cont}
\end{figure*}

The results are shown in \Cref{fig:sm_testres}. For each counterfactual predicted class we report the selection and rejection rate of each loss function across two methods---RelMulti and RelPSI. We can see that for classes with sample size $>200$, the non-zero rejection rates mostly correspond to loss functions $A$ and $B$ which do not include the prototype term.

Note that it is important to perform the test for each class of counterfactual instances separately. The reasoning is the same as for introducing the prototype loss term in the first place as we care about how close in distribution the generated counterfactual instances are to the same class of real instances. If, on the other hand, we run the test across all classes we will not get a reliable measure of this. For example, we noted that running the test across all classes resulted in small rejection rates for loss $A$. This is because the counterfactuals of loss $A$ are generated only using sparsity constraints, thus most counterfactuals resemble the original instance whilst giving a different prediction by the model (akin to adversarial examples). Thus aggregating classes would result in good test results for loss $A$.

\end{document}